%% file: main.tex
\documentclass{article}

\usepackage[final]{nips_2017}

\usepackage[utf8]{inputenc} % allow utf-8 input
\usepackage[T1]{fontenc}    % use 8-bit T1 fonts
\usepackage{hyperref}       % hyperlinks
\usepackage{url}            % simple URL typesetting
\usepackage{booktabs}       % professional-quality tables
\usepackage{amsfonts}       % blackboard math symbols
\usepackage{nicefrac}       % compact symbols for 1/2, etc.
\usepackage{microtype}      % microtypography
\usepackage{amsmath}
\usepackage{amssymb}
\usepackage{graphicx} % more modern
\usepackage{float}
\usepackage{enumitem}
\usepackage{lipsum} 
\usepackage{algorithm}
\usepackage[noend]{algpseudocode}
\newcommand{\kl}[2]{\mathcal{D}_{\textsc{KL}}\big(#1||#2\big)}
\usepackage{subfig} 
\usepackage{color}
\usepackage{bbm}

\newcommand{\remove}[1]{}
\usepackage{tikz}
\usepackage[toc,page]{appendix}
\usepackage{xfrac}
\usepackage{relsize}
\usepackage{float}
\renewcommand{\kl}[2]{\mathcal{D}_{\textsc{KL}}\big(#1||#2\big)}
\usetikzlibrary{fit,positioning,arrows,automata}

\definecolor{right}{HTML}{697c2a}
\definecolor{down}{HTML}{000000}

\title{Disentangling the independently controllable factors of variation by interacting with the world}



\author{
Valentin Thomas$^{*1,2}$
\and
Emmanuel Bengio$^{*3}$
\and 
William Fedus$^{*1}$
\and 
Jules Pondard$^4$
\and
Philippe Beaudoin$^2$
\and
Hugo Larochelle$^{1,5}$
\and
Joelle Pineau$^{3,6}$
\and
Doina Precup$^{3,7}$
\and
Yoshua Bengio$^{1,8}$
}

\begin{document}
\newcommand\blfootnote[1]{%
  \begingroup
  \renewcommand\thefootnote{}\footnote{#1}%
  \addtocounter{footnote}{-1}%
  \endgroup
}
\blfootnote{$^*$Equal contribution, random order, 
$^1$MILA, Universit\'{e} de Montr\'{e}al,
$^2$Element AI,
$^3$McGill University,
$^4$ENS Paris,
$^5$Google Brain,
$^6$Facebook AI Research,
$^7$Google Deepmind,
$^8$CIFAR Senior Fellow
}

\maketitle
% \tableofcontents
\begin{abstract}

    % Optimistic abstract: We propose to expand the framework of Independently Controllable Features to learn fixed-length closed-loop option embeddings. We show that these embeddings learn interesting disentangled representations that uncover aspects of the environment which are \emph{independently controllable}. Additionally, we show that the objective we propose has a close link to the notion of empowerment. We empirically show for the MazeBase domain (hopefully) that the learned representation benefits traditional task-solving agents when used as an auxiliary task, and have interesting properties that allow for some level of model-based prediction and policy inference.\\

  It has been postulated that a good representation is one that disentangles the underlying explanatory factors of variation. However, it remains an open question what kind of training framework could potentially achieve that. Whereas most previous work focuses on the static setting (e.g., with images), we postulate that some of the causal factors could be discovered if the learner is allowed to interact with its environment. The agent can experiment with different actions and observe their effects. More specifically, we hypothesize that some of these factors correspond to aspects of the environment which are independently controllable, i.e., that there exists a policy and a learnable feature for each such aspect of the environment, such that this policy can yield changes in that feature with minimal changes to other features that explain the statistical variations in the observed data. We propose a specific objective function to find such factors, and verify experimentally that it can indeed disentangle independently controllable aspects of the environment without any extrinsic reward signal.
\end{abstract}

\section{Introduction}
% []
% Focus of the Paper: what we want to sell
% \begin{itemize}
% \item Learning disentangled representations
% \item Using an intrinsic motivation goal to do so (+bonus with ELBO)
% \item The selectivity cost is really general, easy to compute and can be used in any "learning in latent space" RL task
% \item With the selectivity cost, it may be possible to move to model-based RL systems that are predicting a non-trivial latent space rather than pixel space.
% \item Link with causality also (will be writing the maths more clearly on that point soon enough)
% \end{itemize}

When solving Reinforcement Learning problems, what separates great results from random policies is often having the right feature representation. Even with function approximation, learning the right features can lead to faster convergence than blindly attempting to solve given problems \citep{jaderberg2016reinforcement}.

The idea that learning good representations is vital for solving most kinds of real-world problems is not new, both in the supervised learning literature \citep{Bengio-2009-book,Goodfellow-et-al-2016}, and in the RL literature \citep{dayan1993improving,precup2000temporal}. An alternate idea is that these representations do not need to be learned explicitly, and that learning can be guided through internal mechanisms of reward, usually called intrinsic motivation \citep{barto2004intrinsically,oudeyer2009intrinsic,salge2013empowerment, varintcontrol}. 

We build on a previously studied \citep{thomas2017independently} mechanism for representation learning that has close ties to intrinsic motivation mechanisms and causality. This mechanism explicitly links the agent's control over its environment to the representation of the environment that is learned by the agent. More specifically, this mechanism's hypothesis is that most of the underlying factors of variation in the environment can be controlled by the agent independently of one another. %For example, in a physical manipulation setting, each distinctly manipulable object is associated to a latent factor, which is controllable by the agent without affecting other factors.

We propose a general and easily computable objective for this mechanism, that can be used in any RL algorithm that uses function approximation to learn a latent space. We show that our mechanism can push a model to learn to disentangle its input in a meaningful way, and learn to represent factors which take multiple actions to change and show that these representations make it possible to perform model-based predictions in the learned latent space, rather than in a low-level input space (e.g. pixels).

\section{Learning disentangled representations}
% Important points:
% \begin{itemize}
% \item Speak about representation learning
% \item Interaction with an environment might help for unsupervised learning
% \item We try to capture controllable factors of variations by trying them
% \item Link with causlity? (maybe we'll cite this paper https://arxiv.org/abs/1605.08179) we can also speak of the "missing signal" as Leon Bottou calls it, the difference is that we not trying to find the causal link between two known variables but rather trying the find the variables that correspond to a known causality
% \item Allows us to do predictions in latent space.
% \end{itemize}

The canonical deep learning framework to learn representations is the autoencoder framework \citep{hinton2006reducing}. There, an encoder $f:S\to H$ and a decoder $g:H \to S$ are trained to minimize the \emph{reconstruction error}, $\|s-g(f(s))\|_2^2$. $H$ is called the latent (or representation) space, and is usually constrained in order to push the autoencoder towards more desirable solutions. For example, imposing that $H\in\mathbb{R}^K, S\in\mathbb{R}^N,\; K\ll N$ pushes $f$ to learn to compress the input; there the bottleneck often forces $f$ to extract the principal factors of variation from $S$. However, this does not necessarily imply that the learned latent space disentangles the different factors of variations. Such a problem motivates the approach presented in this work.

% There is no clear mathematical definition of what \emph{disentangled representations} should be, nonetheless, in practice the impact of such representations is generally that they are easier to visualize and use in transfer or continual learning, and extract some form of semantic meaning from an entangled input space.

%{\color{red} maybe can reduce this section a bit}\\
Other authors have proposed mechanisms to disentangle underlying factors of variation. Many deep generative models, including variational autoencoders~\citep{Kingma+Welling-ICLR2014} \remove{and other descendants of the Helmholtz machine~\citep{Dayan-et-al-1995}}, generative adversarial
networks~\citep{Goodfellow-et-al-NIPS2014} or non-linear versions of ICA~\citep{Dinh-et-al-2014,Hyvarinen+Morioka-NIPS2016} attempt to disentangle the underlying factors of variation by assuming that their joint distribution (marginalizing out the observed $s$) factorizes, i.e., that they are marginally independent.

Here we explore another direction, trying to exploit the ability of a learning agent to act in the world in order to impose a further constraint on the representation. We hypothesize that interactions can be the key to learning how to disentangle the various causal factors of the stream of observations that an agent is faced with, and that such learning can be done in an unsupervised way. %(i.e. with no extrinsic reward from a task).

% NOTE: MB doesn't seem to be ready yet (unreliable, mostly for small environments).
% {\color{red} In recent model-based RL work, most algorithms rely on predictions of future pixel-states \citep{oh2015action} \citep{chiappa2017recurrent}, \citep{weber2017imagination} we show here that our method can perform model-based RL using only a loss defined in the latent space.

% \begin{itemize}
% \item Predicting future \textit{latent} states has many desirable advantages over predicting raw pixel-states.  For instance, raw pixel-states may have non-salient information that is not useful for the task.  A well-learned latent representation may encode only necessary aspects of the environment and may thus be more powerful for model-based RL. 
% \item Of course, one major issue with the prediction of future latent states, rather than future pixel states, is that these may devolve to uninteresting features.  We hypothesize that having an ICF auxillary loss component may help prevent degenerate latent states.  
% \item In our chosen architecture, we are using an autoencoder of the raw state and therefore, we have a functional mapping from our latent representation to pixel-space.  However, this architectural decision is not necessary and ICF may be learned without this.  
% \end{itemize}
% }

% \section{Intrinsic motivation and the causal information}
% Present some works in this area

\section{The selectivity objective}
We consider the classical reinforcement learning setting but in the case where extrinsic rewards are not available. We introduce the notion of \textbf{controllable factors of variation} $\phi \in \mathbb{R}^K$ which are generated from a neural network $\Phi(h, z), z \sim \mathcal{N}(0, 1)^m$ where $h = f(s)$ is the current latent state. The factor $\phi$ represents an embedding of a policy $\pi_\phi$ whose goal is to realize the variation $\phi$ in the environment.

To discover meaningful factors of variation $\phi$ and their associated policies $\pi_\phi$, we consider the following general quantity $\mathcal{S}$ which we refer to as selectivity and that is used as a reward signal for $\pi_\phi$:
\begin{equation}
    \mathcal{S}(h, \phi) = \mathbb{E}  \left[\log \frac{A(h', h, \phi)}{\mathbb{E}_{p(\varphi|h)}[A(h', h, \varphi)]} \;\big|\; \; {s' \sim \mathcal{P}^{\pi_\phi}_{ss'}}\right]
      \label{eq:general-selectivity}
\end{equation}

%Here, $h$ is the initial latent state and $h'$ is the final latent state. Given an encoder $f$, we therefore have $h = f(s), h' = f(s')$.
Here $h=f(s)$ is the encoded initial state before executing $\pi_\phi$ and   $h' = f(s')$ is the encoded terminal state. $\phi$ and $\varphi$ represent factors of variation a \emph{factor}.
$A(h', h, \phi)$ should be understood as a score describing how close $\phi$ is to the variation it caused in $(h', h)$. For example in the experiments of section 4.1, we choose $A$ to be a gaussian kernel between $h'-h$ and $\phi$, while in the experiments of section 4.2, we choose $A(h', h, \phi) = \max\{0,\langle h'-h, \phi \rangle\}$.
The intuition behind these objectives is that in expectation, a factor $\phi$ should be close to the variation it caused $(h', h)$ when following $\pi_\phi$ compared to other factors $\varphi$ that could have been sampled and followed thus encouraging \textbf{independence} within the factors.

Conditioned on a scene representation $h$, a distribution of policies are feasible. Samples from this distribution represent ways to modify the scene and thus may trigger an internal selectivity reward signal. For instance, $h$ might represent a room with objects such as a light switch. $\phi = \phi(h,z)$ can be thought of as the distributed representation for the ``name'' of an underlying factor, to which is associated a policy and a value.  In this setting, the light in a room could be a factor that could be either on or off. It could be associated with a policy to turn it on, and a binary value referring to its state, called an attribute or a feature value.% \textit{turn on the light} would be relevant and thus likely to be sampled, while \textit{go skydiving} would not. To every $\phi$, we associate a relevant attribute $A(h, \phi)$ of the environment and a policy $\pi_\phi$ whose goal is to maximally impact attribute $A(h, \phi)$ compared to others. In our example, an attribute likely to be modified would be the brightness level of the room.
We wish to jointly learn the policy $\pi_\phi(\cdot|s)$ that modifies the scene, so as to control the corresponding value of the attribute in the scene, whose variation is computed by a scoring function $A(h', h, \phi) \in \mathbb{R}$. In order to get a distribution of such embeddings, we compute $\phi(h,z)$ as a function of $h$ and some random noise $z$.

% In this scenario, one strategy to determine whether some selected attribute variation $A(h' - h,\phi)$ evolves \emph{independently} from other attributes variations is to compare its value (in expectation over the policy actions) to the values obtained with other $\phi'$ factors. We thus compute the following selectivity that acts as an intrinsic reward signal, generalizing \eqref{eq:sel}:}

% $\Delta(h',h;\alpha)$ represents the quantity of change from $h$ to $h'$ of the encoding of $\alpha$ in $h$. $\pi_\alpha$ represents a unique policy, associated with each unique $\alpha$, and $\mathcal{P}^\alpha_{ss'}$ represents the (not necessarily single-step) transition probability from $s$ to $s'$ when following $\pi_\alpha$.

The goal of a selectivity-maximizing model is to find the density of factors $p(\phi |h)$, the latent representation $h$, as well as the policies $\pi_\phi$ that maximize $\mathbb{E}_{p(\phi|h)} [\mathcal{S}(h, \phi)]$.

% \iscopy{Is this helping with intuition? Let's maybe keep the blue part rather}\\
% {\color{red}
% Let's illustrate this in the discrete case, where $\phi^{(i)}$ is a one-hot vector with entry $i$ equal to one and $p(\Phi = \phi^{(i)} | h) = \frac{1}{K}$. $\phi^{(i)}$ selects the $i$th dimension of $h\in \mathbb{R}^K$, or $h_i$. There we learn $K$ latent factors and $K$ policies $\pi_i$. This objective expresses the fact that we want to find a representation where at each transition $(s_t,a_t,s_{t+1})$, only one coordinate changes, say $h_k$, and it changes due to sampling from $\pi_k(a_t|s_t)$. That is, following $\pi_k$ should only change one "independently controllable factor" of the state space.

% In a more general setting, we jointly learn the encoder $f$ the distribution $p(\Phi=\phi | H = h)$ which we can sample from, and the parameterized policies $\pi_\phi(a|h)$.}

\begin{figure}
\centering
\subfloat[]{\input{embed_fig}}
\caption{The computational model of our architecture. $s_t$ is the first state, from its encoding $h_t$ and a noise distribution $z$, $\phi$ is generated. $\phi$ is used to compute the policy $\pi_\phi$, which is used to act in the world. The sequence $h_t,h_{t'}$ is used to update our model through the selectivity loss, as well as an optional autoencoder loss on $h_t$.}
\end{figure}
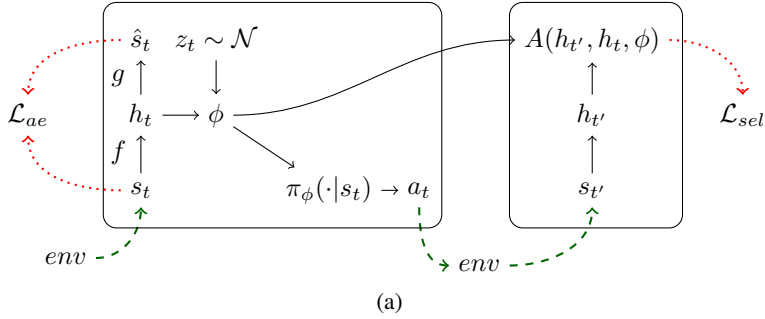
% it is reasonable to expect the aforementioned behaviour from our latent space. That is, we should find independently controllable factors.

\subsection{Link with mutual information and causality}
% Important points:
% \begin{itemize}
% \item Maximizing an objective defined directly in latent space
% \item Tackles the issue of causality, empowerment, model-based RL by proposing a suitable objective function
% \item Still different from empowerment as finding the "empowering" policy is not our ultimate goal
% \end{itemize}
The selectivity objective, while intuitive, can also be related to information theoretical quantities defined in the latent space.
From \citep{donsker1975asymptotic,ruderman2012tighter} we have
$\kl{p}{q} = \sup_{A\in \mathcal{L}^\infty(q)} \mathbb{E}_p [ \log A] - \log \mathbb{E}_q [A]$.
Applying this equality to the mutual information $\mathcal{I}_p(\phi, h' | h) = \mathbb{E}_{p(h'|h)}\big[ \kl{p(\phi|h', h)}{p(\phi|h)}\big]$
gives
$$\mathcal{I}_p(\phi, h' | h) \ge \sup_\theta  \mathbb{E}_{p(\phi|h)} \big[ \mathcal{S}(h, \phi)\big]$$
where $\theta$ is the set of weights shared by the factor generator, the policy network and the encoder.

Thus, our total objective along entire trajectories is a lower bound on the causal \citep{ziebart2010modeling} or directed \citep{massey1990causality} information $\mathcal{I}_p(\phi \mapsto h) = \sum_t \mathcal{I}_p(\phi_{1:t}, h_t|h_{t-1})$ which is a measure of the \textbf{causality} the process $\phi$ exercises on the process $h$. See Appendix \ref{appendix:bound} for details.

% \subsection{Parametrizing complex policies}
% Speak about long term policies the dot product and the amplitude

\section{Experiments}
\remove{\color{red} So do we
\begin{enumerate}
\item Keep only this experiment (we wouldn't have to worry much about the space but that is light on the experiments)
\item Add the one with planning/policy inference?~\ref{sec:exp-mb-ppi} (small example from the paper)
\item Add the one with factors that seem to like to toggle switch in multistep~\ref{sec:multistep} (though not the behaviour we would really expect)
\item Add the discrete one?~\ref{sec:sel-only}
\end{enumerate}
Seems the experiments is working, doing a figure....}
We use MazeBase~\citep{sukhbaatar2015mazebase} to assess the performance of our approach. We do not aim to solve the game.
In this setting, the agent (a red circle) can move in a small environment ($64 \times 64$ pixels) and perform the actions \texttt{down, left, right, up}. The agent can go anywhere except on the orange blocks.

% \subsection{Policy embedding in complex env}
% So we do discover switches (one-step or multistep) reliably only if we use a directed selectivy, that is $|\langle h'-h, \phi \rangle|$ for example, but is not reliable with directed selectivity such as $1 + \frac{\langle h'-h, \phi \rangle}{||h'-h|| \cdot ||\phi||}$ or $\texttt{ReLU}(\langle h'-h, \phi \rangle)$

% TODO: alternative figure to above
% Cross correlation $corr(A(h, phi), A*)$ Cannot really do that as $\phi$ also depends on $h$...

\subsection{Learned representations}
\begin{figure}
\centering
\subfloat[]{\includegraphics[width=.3\linewidth]{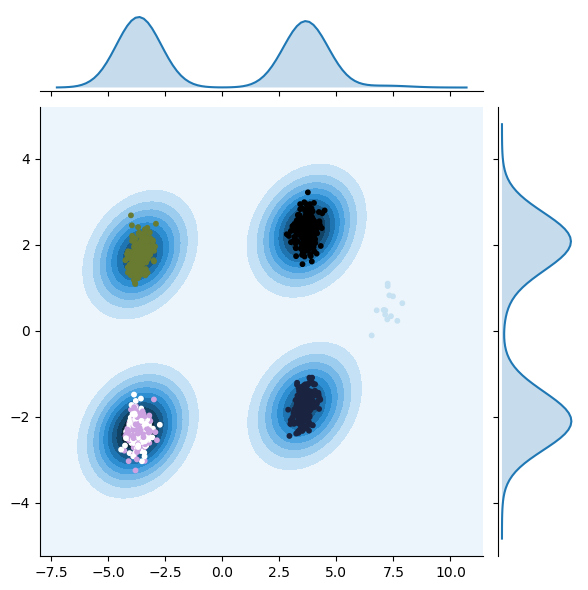}}
\hspace{1cm}%\hfill
%\subfloat[]{\includegraphics[width=.3\linewidth]{./images/pdoth_tsne_p.png}}
\subfloat[]{\includegraphics[width=.3\linewidth]{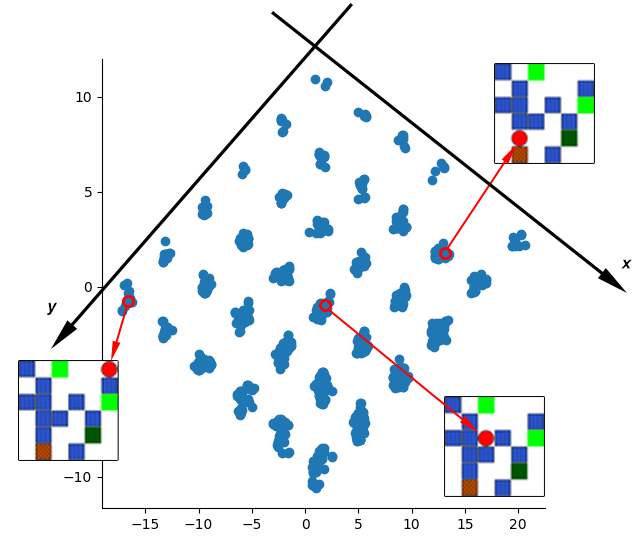}} %trim={0px 0 0px 0},clip
% \hfill
% \subfloat[]{\includegraphics[width=.34\linewidth]{./images/25000_kde_phi.png}}

\caption[test]{(a) Sampling of $1000$ variations $h' - h$ and its kernel density estimation encountered when sampling random controllable factors $\phi$. We observe that our algorithm disentangles these representations on $4$ main modes, each corresponding to the action that was actually taken by the agent.\protect\footnotemark
\, (b) The disentangled structure in the latent space. The $x$ and $y$ axis are disentangled such that we can recover the $x$ and $y$ position of the agent in any observation $s$ simply by looking at its latent encoding $h = f(s)$. The missing point on this grid is the only position the agent cannot reach as it lies on an orange block.}
\label{fig:dis_space}
\end{figure}
\footnotetext{pink and white for \texttt{up}, light blue for \texttt{down+left}, green for \texttt{right}, purple black \texttt{down} and night blue for \texttt{left}.}

After jointly training the reconstruction and selectivity losses, our algorithm disentangles four directed factors of variations as seen in Figure~\ref{fig:dis_space}: $\pm x$-position and $\pm y$-position of the agent. For visualization purposes we chose the bottleneck of the autoencoder to be of size $K = 2$.
To complicate the disentanglement task, we added the redundant action \texttt{up} as well as the action \texttt{down+left} in this experiment.

The disentanglement appears clearly as the latent features corresponding to the $x$ and $y$ position are orthogonal in the latent space. Moreover, we notice that our algorithm assigns both actions \texttt{up} (white and pink dots in Figure~\ref{fig:dis_space}.a) to the same feature. It also does not create a significant mode for the feature corresponding to the action \texttt{down+left} (light blue dots in Figure~\ref{fig:dis_space}.a) as this feature is already explained by features \texttt{down} and \texttt{left}.

% \subsubsection{Model based predictions}
% Show how adding a MB term can allow us to do prediction in latent space only
% It works but only on "discovered" $\phi$ which is not enough...
% \subsubsection{Policy inference}
% If we have access to the posterior of $q$, $q(\phi|h', h)$ we can do policy inference

% \subsection{Using factors for (model based?) task solving}

\subsection{Multistep embedding of policies}
In this experiment, $\phi$ are embeddings of $3$-steps policies $\pi_\phi$. We add a model-based loss $\mathcal{L}_{MB}=||h_{t+3} - T_\theta(h_t, \phi)||^2$ defined only in the latent space, and jointly train a decoder alongside with the encoder. Notice that we never train our model-based cost at pixel level.
While we currently suffer from mode collapsing of some factors of variations, we show that we are successfully able to do predictions in latent space, reconstruct the latent prediction with the decoder, and that our factor space disentangles several types of variations.

\begin{figure}
\centering
%\subfloat[]{\raisebox{-0.5\height}{\includegraphics[width=.25\linewidth]{./images/square_traj.png}}} %trim={0px 0 0px 0},clip
%\subfloat[]{\raisebox{-0.5\height}{\includegraphics[width=.45\linewidth]{./images/phi_space.png}}}
\subfloat[]{{\includegraphics[width=.25\linewidth]{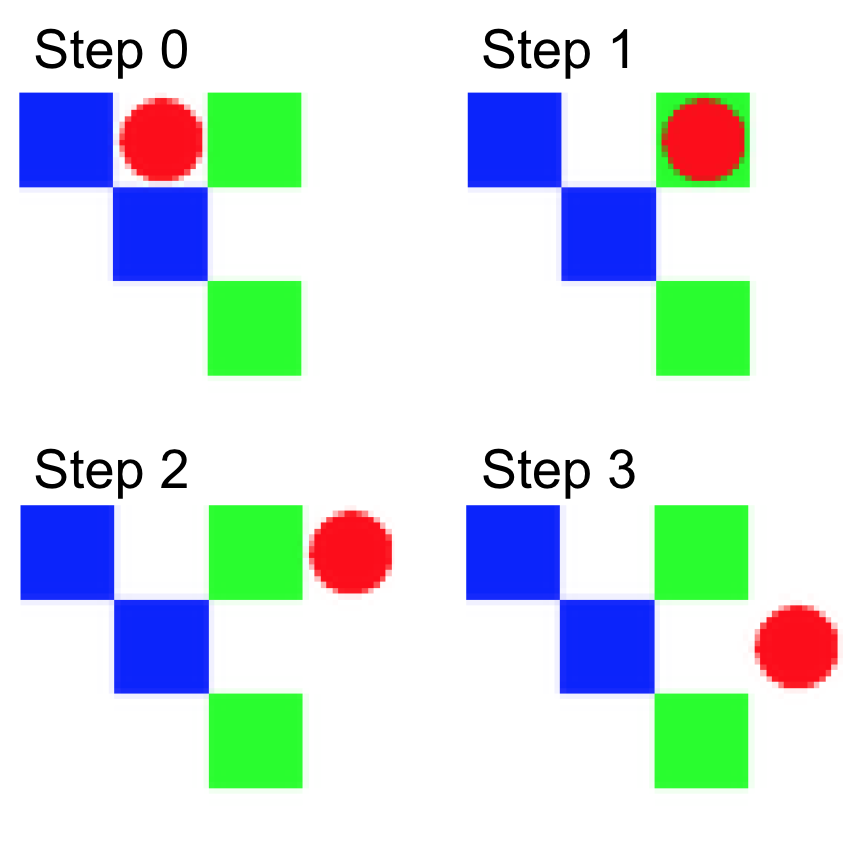}}} %trim={0px 0 0px 0},clip
\subfloat[]{{\includegraphics[width=.45\linewidth]{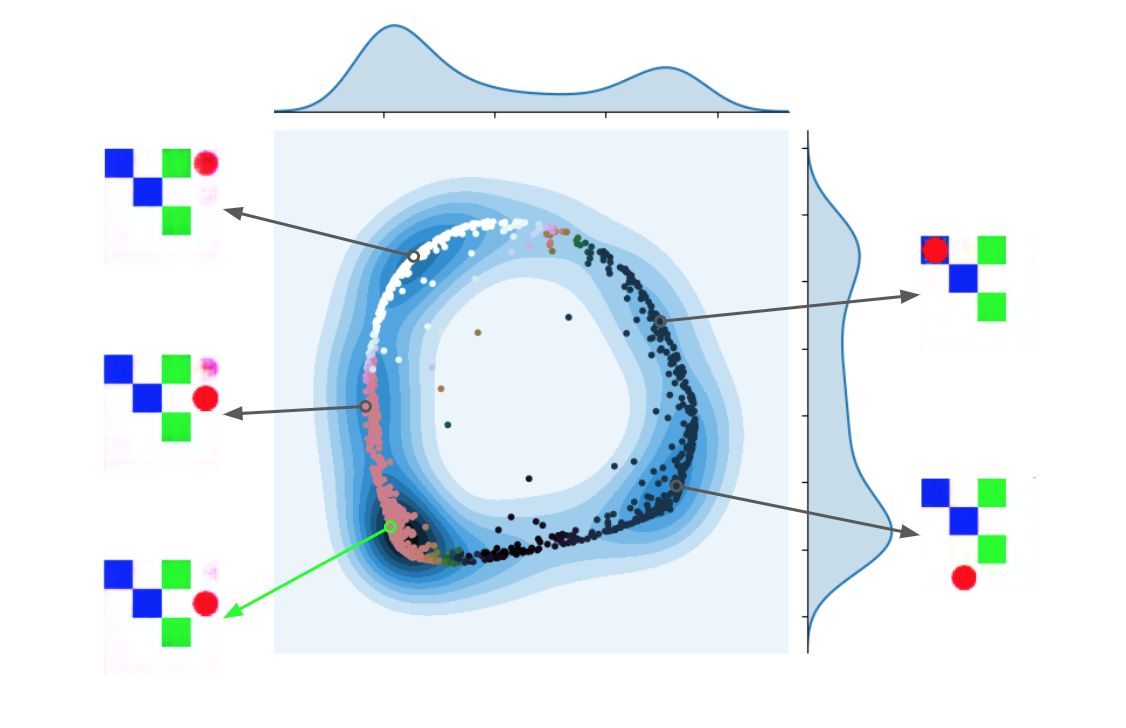}}}
%\hfill
%\subfloat[]{\includegraphics[width=.3\linewidth]{./images/pdoth_tsne_p.png}}

% \hfill
% \subfloat[]{\includegraphics[width=.34\linewidth]{./images/25000_kde_phi.png}}
\caption{(a) The actual 3-step trajectory done by the agent. (b) PCA view of the space $\phi(h_0, z), z \sim \mathcal{N}(0,1)$. Each arrow points to the reconstruction of the prediction $T_\theta(h_0, \phi)$ made by different $\phi$. The $\phi$ at the start of the green arrow is the one used by the policy in (a). Notice how its prediction accurately predicts the actual final state.}
\end{figure}

\section{Conclusion, success and limitations}
Pushing representations to model independently controllable features currently yields some encouraging success. Visualizing our features clearly shows the different controllable aspects of simple environments, yet, our learning algorithm is unstable. What seems to be the strength of our approach could also be its weakness, as the independence prior forces a very strict separation of concerns in the learned representation, and should maybe be relaxed.

Some sources of instability also seem to slow our progress: learning a conditional distribution on controllable aspects that often collapses to fewer modes than desired, learning stochastic policies that often optimistically converge to a single action, tuning many hyperparameters due to the multiple parts of our model. Nonetheless, we are hopeful in the steps that we are now taking. Disentangling happens, but understanding our optimization process as well as our current objective function will be key to further progress. 
% Acknowledgements should only appear in the accepted version. 
%\section*{Acknowledgements} 
%\vspace*{-10pt}

\bibliography{main}
\bibliographystyle{plainnat}

\appendix
\input{appendix.tex}

\end{document}

%% file: embed_fig.tex
\begin{tikzpicture}

\def \pA {0}
\def \pB {6}
\def \pmid {2.5}

\node[] (st) at (\pA,0) {$s_t$};
\node[] (ht) at (\pA,1) {$h_t$};
\node (st_rec) at (\pA,2) {$\hat{s}_t$};

\draw[->,draw=black] (st) to (ht);
\draw[->,draw=black] (ht) to (st_rec);
\node at (-.3,1.5) {$g$};
\node at (-.3,0.5) {$f$};

\node (env) at (-1,-0.9) {$env$};
\draw[->,draw=black!60!green,dashed,thick] (env) to[out=0,in=-90] (st);

\node[] (st1) at (\pB,0) {$s_{t'}$};
\node[] (ht1) at (\pB,1) {$h_{t'}$};
%\node (st1_rec) at (\pB,2) {$\hat{s}_{t+1}$};
\draw[->,draw=black] (st1) to (ht1);
%\draw[->,draw=black] (ht1) to (st1_rec);

\node[] (pi0) at (\pmid,0) {$\pi_\phi(\cdot|s_t)$};
\node[] (phi0) at (\pmid-1.5,1) {$\phi$};
\node[] (z0) at (\pmid-1.5,2) {$z_t \sim \mathcal{N}$};
% \node[] (A0) at (\pmid,2) {$A(h_t,\phi)$};
\node[] (A1) at (\pB,2) {$A(h_{t'}, h_t,\phi)$};
\draw[->,draw=black] (z0) to (phi0);
\draw[->,draw=black] (ht) to (phi0);
\draw[->,draw=black] (phi0) to (pi0);
% \draw[->,draw=black] (phi0) to (A1);
\draw[->,draw=black] (phi0) to[in=180,out=0] (A1);
\draw[->,draw=black] (ht1) to (A1);

\begin{scope}[]

\end{scope}
\node[] (aT) at (\pmid+1.2,-0) {$a_t$};
\node[] (env1) at (\pmid+2,-1) {$env$};
\draw[->] (pi0) to (aT);
\draw[->,draw=black!60!green,dashed,thick] (aT) to[out=-90,in=180] (env1);
\draw[->,draw=black!60!green,dashed,thick] (env1) to[out=0,in=-90] (st1);
% \draw[->,draw=black!60!green,dashed,thick] (env1) to[out=0,in=90] (pi0);

\node[] (lossAE) at (\pA-1.5,1) {$\mathcal{L}_{ae}$};
\draw[->,draw=red,dotted,thick] (st) to[out=180,in=-90] (lossAE);
\draw[->,draw=red,dotted,thick] (st_rec) to[out=180,in=90] (lossAE);

\node[] (lossSel) at (\pB+2,1) {$\mathcal{L}_{sel}$};
% \draw[->,draw=red,dotted,thick] (A0) to[out=5,in=90] (lossSel);
\draw[->,draw=red,dotted,thick] (A1) to[out=0,in=90] (lossSel);

\draw
  {[rounded corners=5pt](-0.5,-0.5)  --
  (\pmid+1.5,-0.5)  -- 
  (\pmid+1.5,2.5) --
  (-0.5,2.5) --
  cycle};
\draw
  {[rounded corners=5pt](-1.1+\pB,-0.5)  --
  (1.2+\pB,-0.5)  -- 
  (1.2+\pB,2.5) --
  (-1.1+\pB,2.5) --
  cycle};
\end{tikzpicture}

%% file: appendix.tex
\section{Additional details}
\subsection{Architecture}
Our architecture is as follows: the encoder, mapping the raw pixel state to a latent representation, is a 4-layer convolutional neural network with batch normalization \citep{ioffe2015batch} and leaky ReLU activations. The decoder uses the transposed architecture with ReLU activations. The noise $z$ is sampled from a 2-dimensional gaussian distribution and both the generator $\Phi(h,z)$ and the policy $\pi(h,\phi)$ are neural networks consisting of 2 fully-connected layers. In practice, a minibatch of $n = 256$ or $1024$ vectors $\phi_1, \dots, \phi_{n}$ is sampled at each step. The agent randomly choses one $\phi = \phi_{behavior}$ and samples actions from its policy $a \sim \pi(h, \phi_{behavior})$. Our model parameters are then updated using policy gradient with the REINFORCE estimator and a state-dependent baseline and importance sampling. For each selectivity reward, the term $\mathbb{E}_{\phi'}[  A(h', h,\phi')]$ is estimated as $\tfrac{1}{n} \sum_{i = 1}^n A(h', h,\phi_i)$.

In practice, we don't use concatenation of vectors when feeding two vectors as input for a network (like $(h, z)$ for the factor generator or $(h, \phi)$ for the policy).
For vectors $a,b \in \mathbb{R}^{n_a \times n_b}$. We use a bilinear operation $bil(a, b) = (a_i * b_j)_{i \in [[n_a]], j \in [[n_b]]}$ as in \cite{florensa2017stochastic}.  We observe the bilinear integrated input to more strongly enforce dependence on both vectors; in contrast, our models often ignored one input when using a simple concatenation.

Through our research, we experiment with different outputs for our generator $\Phi(h,z)$.  We explored embedding the $\phi$-vectors into a hypercube, a hypersphere, a simplex and also a simplex multiplied by the output of a $tanh(\cdot)$ operation on a scalar.

\subsection{First experiment}
In the first experiment, figure~\ref{fig:dis_space}, we used a gaussian similarity kernel i.e $A(h', h, \phi) = \exp(- \frac{||h' - (h+\phi)||^2}{2 \sigma^2})$ with $\sigma = \sqrt{dim(h)}$.
In this experiment only, for clarity of the figure, we only allowed permissible actions in the environment (no no-op action).

% Our attribute selector $A(dh, \phi)$ is a gaussian kernel.

\section{Additional Figures}
\subsection{Discrete simple case}
\label{sec:sel-only}
Here we consider the case where we learn a latent space $H$ of size $K$, with $K$ factors corresponding to the coordinates of $h$ ($h_i,\; i\in[k])$, and learn $K$ separately parameterized policies $\pi_i(a|h),\;i\in[k]$. We train our model with the selectivity objective, but no autoencoder loss, and find that we correctly recover independently controllable features on a simple environment.
{%We also find experimentally that training discrete independently controllable features without training the autoencoder objective correctly recovers ground truth features and their associated control policies. 
Albeit slower than when jointly training an autoencoder, this shows that the objective we propose is strong enough to provide a learning signal for discovering a disentangled latent representation.
}

\begin{figure}[H]
\centering
\includegraphics[width=\linewidth]{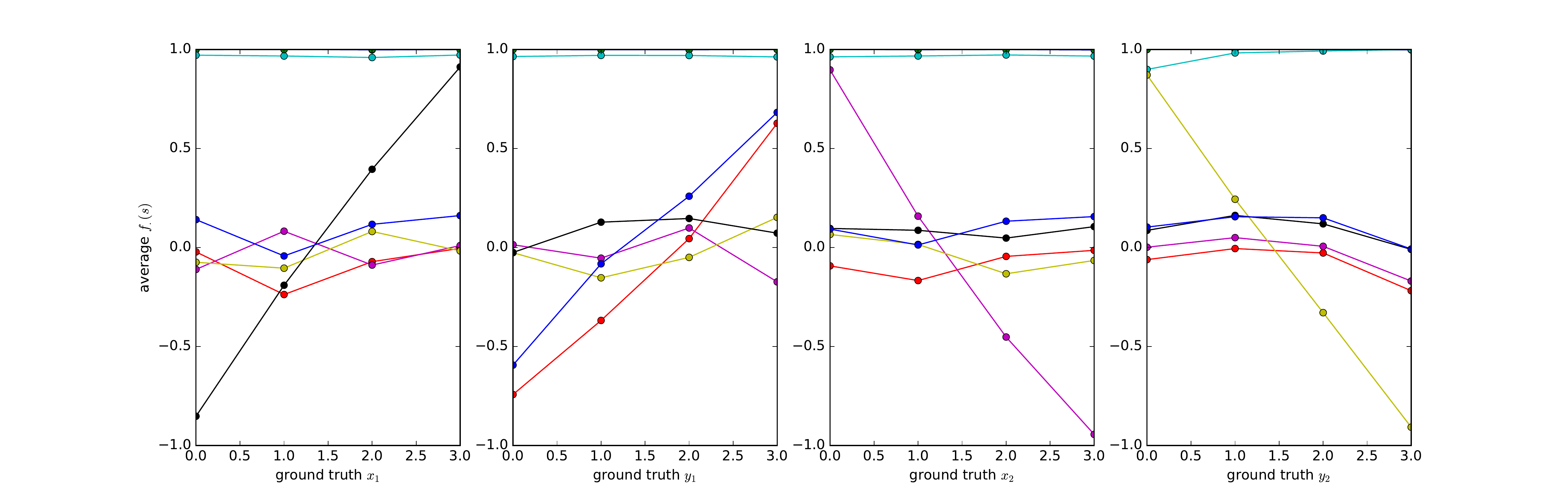}
\caption{{%Each curve represents a latent feature $f_k$ as a function of some ground truth. Here there are 4 ground truths, the $x$ and $y$ position of each of the 2 MNIST digits.
In a gridworld environment with 2 objects (in this case 2 MNIST digits), we know there are 4 underlying features, the $(x_i,y_i)$ position of each digit $i$. Here each of the four plots represents the evolution of the $f_k$'s as a function of their underlying feature, from left to right $x_1$, $y_1$, $x_2$, $y_2$. We see that for each of them, at least one $f_k$ recovers it almost linearly, from the raw pixels only.}
}
\label{fig:gridworld-only-sel}
\end{figure}

{We train such a model on a gridworld MNIST environment, where there are two MNIST digits \nocite{lecun1998mnist}. The two digits can be moved on the grid via 4 directional actions (so there are 8 actions total), the first digit is always odd and the second digit always even, so they are distiguishable. In Figure \ref{fig:gridworld-only-sel} we plot each latent feature $h_k$ as a curve, as a function of each ground truth. For example we see that the black feature recovers $+x_1$, the horizontal position of the first digit, or that the purple feature recovers $-y_2$, the vertical position of the second digit.}

\subsection{Planning and policy inference example in 1-step}
\label{sec:exp-mb-ppi}

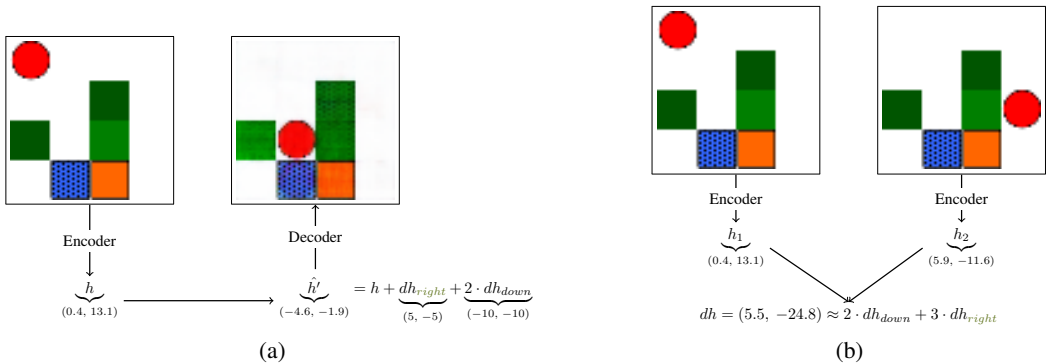
\begin{figure}[H]
\centering
\subfloat[]{\input{prediction.tex}}
\hfill
%\subfloat[]{\includegraphics[width=.3\linewidth]{./images/pdoth_tsne_p.png}}
\subfloat[]{\input{recovering.tex}} %trim={0px 0 0px 0},clip
\caption{(a) Predicting the effect of a cause on Mazebase. The leftmost image is the visual input of the environment, where the agent is the round circle, and the switch states are represented by shades of green. After the training, we are able to distinguish one cluster per $dh$ (Figure \ref{fig:dis_space}), that is to say per variation obtained after performing an action, independently from the position $h$. Therefore, we are able to move the agent just by adding the corresponding $dh$ to our latent representation $h$. The second image is just the reconstruction obtained by feeding the resulting $h'$ into the decoder. (b) Given a starting state and a goal state, we are able to decompose the difference of the two representations $dh$ into a (non-directed) sequence of movements.}
\label{fig:prediction_recovering}
\end{figure}

This disentangled structure could be used to address many challenging issues in reinforcement learning. We give two examples in figure~\ref{fig:prediction_recovering}: 
\begin{itemize}
\item Model-based predictions: Given an initial state, $s_0$, and an action sequence $a_{\{0:T-1\}}$, %the agent performed, our goal is to predict the resulting state $s_T$.
we want to predict the resulting state $s_T$.

\item A simplified deterministic policy inference problem: Given an initial state $s_{start}$ and a terminal state $s_{goal}$, we aim to find a suitable action sequence $a_{\{0:T-1\}}$ such that $s_{goal}$ can be reached from $s_{start}$ by following it.
\end{itemize}
Because of the $tanh$ activation on the last layer of $\Phi(h, z)$, the different factors of variation $dh = h' - h$ are placed on the vertices of a hypercube of dimension $K$, and we can think of the the policy inference problem as finding a path in that simpler space, where the starting point is $h_{start}$ and the goal is $h_{goal}$. We believe this could prove to be a much easier problem to solve.

\subsection{Multistep Example}
\label{sec:multistep}
\begin{figure}[H]
\centering
\subfloat[]{\includegraphics[width=.8\linewidth]{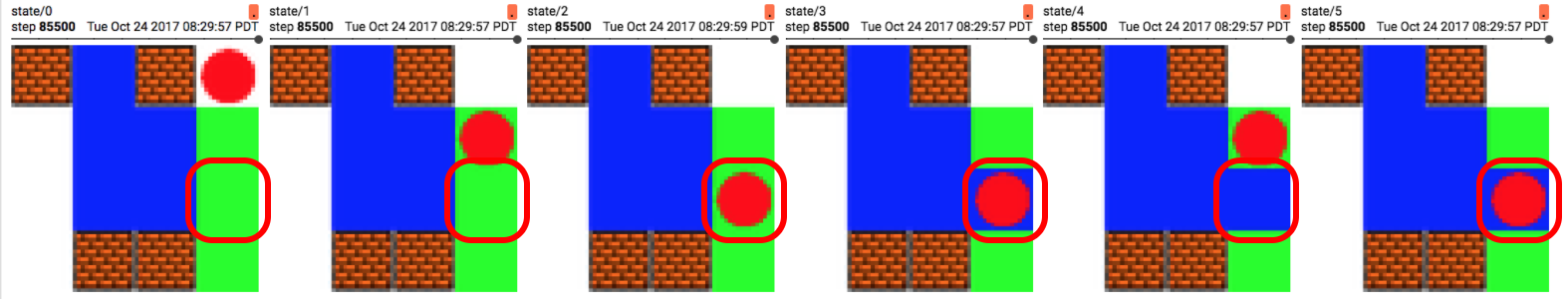}}
\newline
\subfloat[]{\includegraphics[width=.8\linewidth]{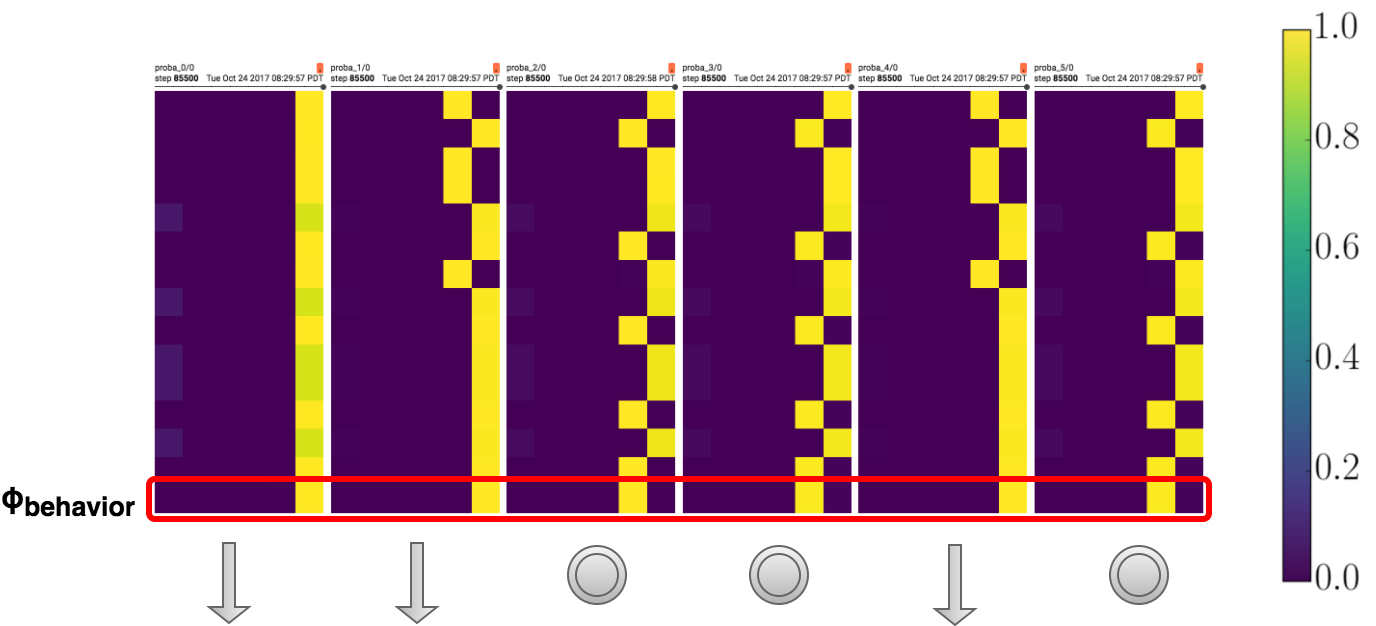}} 
\caption{(a) Mazebase environment over five time-steps.  Here the red dot denotes the position of the agent.  The $\phi_{behavior}$ governing the agent's policy appears to control toggling the switch indicated by the red rounded box. 
(b) Visualization of the policies instantiated by different $\phi$s.  Each box represents the probability distribution of the policies at that time step.  Each row is generated by a different $\phi$ and each column corresponds to an action (up, left, pass, right, toggle, down) in order.  The boxed column shows the $\phi_{behavior}$. The symbols below each box represent the most-probable action for the behavioral policy, where the grey circle indicates toggling the switch.}
\label{fig:multistep_toggle}
\end{figure}

We demonstrate an instance of ICF operating in a 4$\times$4 Mazebase enviroment over five time steps in Figure \ref{fig:multistep_toggle}. We consistently witness a failure of mode collapse in our generator $\Phi$ and therefore the generator only produces a subset of all possible $\phi$-variations.  In Figure \ref{fig:multistep_toggle}, we observe the $\phi$ governing the agent's policy $\pi_{\phi}$ appears to correspond to moving two positions down and then to repeatedly toggle the switch.  A random action due to $\epsilon$-greedy led to the agent moving up and off the switch at time step-4. This perturbation is corrected by the policy $\pi_{\phi}$ by moving down in order to return to toggling the relevant switch.

\section{Variational bound and the selectivity}
\label{appendix:bound}
Let us call $p(h_{t+1} | \phi_{t+1}, h_t) =  \mathcal{P}^\phi_{h' ,h}$ the probability distribution over final hidden states starting from $h$ and using the policy parametrized by the embedding $\phi$.

$p(h_{t+1} | \phi_{t+1}, h_t) = \Pi_{k=1}^{K}  \pi_{\phi_{t+1}}( a_{t+\frac{k-1}{K}} | h_{t+\frac{k-1}{K}}) p_{env}(s_{t+\frac{k}{K}} | a_{t+\frac{k-1}{K}}, s_{t+\frac{k-1}{K}})$.
where $p_{env}$ is the transition probability of the environment.
% $p(\bullet | \phi, h)$ 

For simplicity, let's refer to $h_t$ as $h$, $h_{t+1}$ as $h'$ and $\phi_{t+1}$ as $\phi$.
\subsection{Lower bound on the mutual information}
The bound $$\mathcal{I}_p(\phi, h' | h) \ge \sup_\theta  \mathbb{E}_{p(\phi|h)} \big[ \mathcal{S}(h, \phi)\big]$$ can be proven by using Donsker-Varadhan variational representation of the KL divergence \citep{donsker1975asymptotic,ruderman2012tighter}:

$$\kl{p}{q} = \sup_{T \in \mathcal{L}^\infty(q)} \mathbb{E}_p \big[ T  \big] - \log \mathbb{E}_q \big[ e^T \big]$$

For $A = e^T$ and using the identity $\mathcal{I}_p(X, Y) = \mathbb{E}_{p(y)} \big[ \kl{p(x|y)}{p(x)}\big]$ with $X=\phi|h$ and $Y=h'|h$, we have:

\begin{eqnarray*}
\mathcal{I}_p(\phi, h' | h) &=& \mathbb{E}_{h'|h} \sup_A \mathbb{E}_{\phi|h, h'} \big[ \log A(h', h, \phi) \big] - \log \mathbb{E}_{\varphi|h} \big[ A(h', h, \varphi) \big]\\
&=& \mathbb{E}_{h'|h} \sup_A \mathbb{E}_{\phi|h, h'} \big[ \log \frac{ A(h', h, \phi)}{\mathbb{E}_{\varphi|h} \big[ A(h', h, \varphi)\big]}\big]\\
&\ge&  \sup_A \mathbb{E}_{\phi|h}\mathbb{E}_{h'|\phi, h} \big[ \log \frac{ A(h', h, \phi)}{\mathbb{E}_{\varphi|h} \big[ A(h', h, \varphi)\big]}\big]\\
&\ge&  \sup_\theta \mathbb{E}_{\phi|h}\mathbb{E}_{h'|\phi, h} \big[ \log \frac{ A(h', h, \phi; \theta)}{\mathbb{E}_{\varphi|h} \big[ A(h', h, \varphi; \theta)\big]}\big]\\
\end{eqnarray*}
for parametric $A$ functions.

% In the case where $A(h', h, \phi)$ is actually is probability density $q(h' | h, \phi)$ (or any unnormalized density such that the normalization factor does not depend on $\phi$) we have:
% \begin{eqnarray*}
% \mathcal{S}(\phi, h) &=&\mathbb{E}_{h' \sim p
%     (h' | \phi, h)}  \log \frac{q(h' | \phi, h)}{\mathbb{E}_{\varphi | h} q(h' |
%     \varphi,
% h)}\\
% &=& \mathbb{E}_{\phi | h} \big[ {\kl{p(h'|\phi, h)}{q(h'|h)} -
% \kl{p(h'|\phi, h)}{q(h'|\phi, h)}} \big]\\
% &=& \mathbb{E}_{\phi | h} \big[ \kl{p(h'|\phi, h)}{q(h'|h)}\big] -
% \kl{p(h'| h)}{q(h'|h)} - \mathbb{E}_{p(h'|h)} \big[ \kl{p(\phi| h', h)}{q(\phi |h', h)}\big] \\
% &=& \mathbb{E}_{\phi | h} \big[ \kl{p(h'|\phi, h)}{p(h'|h)}\big]- \mathbb{E}_{p(h'|h)} \big[ \kl{p(\phi| h', h)}{q(\phi |h', h)}\big] \\
% &=& \mathcal{I}^{p} (\phi , h'|h) - \mathbb{E}_{p(h'|h)} \big[ \kl{p(\phi| h', h)}{q(\phi |h', h)}\big] \\
% % &=& \mathcal{I}^{p}(\phi \mapsto h')- \mathbb{E}_{p(h'|h)} \big[ \kl{p(\phi| h', h)}{q(\phi |h', h)}\big] \\
% \end{eqnarray*}
% where we used that fact that $p(\phi|h) = q(\phi|h)$ and by design we only sample $\phi$s from one generator.

% Thus, the gap between the selectivity and the mutual information is the KL divergence of the two posterior distributions.

As we sample the factors $\phi$ uniformly, our total objective is then a lower bound on $\sum_t \mathcal{I}(\phi_t, h_t | h_{t-1})$ which corresponds here to the \emph{directed information} \citep{massey1990causality} \cite{ziebart2010modeling} as $\phi_t$ is sampled independently from $\phi_{1:t-1}$.

\section{Additional information on the training}

In our experiments, we use the selectivity objective, an autoencoding loss and an entropy regularization loss $\mathcal{H}(\pi_\phi)$ for each of the policies $\pi_\phi$. Furthermore, in experiment 4.2 we added the model-based cost $||h' - T(h, \phi)||^2$ with $T$ a learned two layer fully connected neural network.

The selectivity is used to update the parameters of the encoder, factor generator and policy networks.
We use the following equation for computing the gradients

$$\nabla_\theta \mathbb{E}_{\pi_\theta} \big[ f_\theta \big] =  \mathbb{E}_{\pi_\theta} \big[ \nabla_\theta f_\theta + f_\theta \nabla_\theta \log \pi_\theta \big]$$ 
We also use a state dependent baseline $V$ as a control variate to reduce the variance of the REINFORCE estimator.

Furthermore, to be able to train the factor generator efficiently, we train all $\phi$ sampled in a mini-batch (of size $1024$) by importance sampling on the probability ratio of the trajectory under each $\phi$

%% file: prediction.tex
\scalebox{.6}{
\begin{tikzpicture}
\node[inner sep=0pt] (im0) at (0,0)
    {\fbox{\includegraphics[width=.25\textwidth]{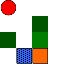}}};
\node[inner sep=0pt] (im1) at (5,0)
    {\fbox{\includegraphics[width=.25\textwidth]{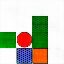}}};

\node[] (ht) at (0,-4) {$\underbrace{h}_{(0.4,\ 13.1)}$};
\node[] (ht1) at (7,-4) {$\underbrace{\hat{h'}}_{(-4.6,\ -1.9)} = h + \underbrace{dh_{{\color{right} {\color{right} right}}}}_{(5,\ -5)} + \underbrace{2\cdot dh_{{\color{down} down}}}_{(-10,\ -10)}$}; 
    \draw[->,thick] (im0.south) -- (ht.north)
    node[midway,fill=white] {Encoder};
    \draw[->,thick] (5, -3.3) -- (im1.south)
    node[midway,fill=white] {Decoder};
    
    \draw[->,thick] (ht.east) -- (ht1.west);
    
\end{tikzpicture}}

%% file: recovering.tex
\scalebox{.6}{
\begin{tikzpicture}
\node[inner sep=0pt] (im0) at (0,0)
    {\fbox{\includegraphics[width=.25\textwidth]{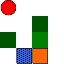}}};
\node[inner sep=0pt] (im1) at (5,0)
    {\fbox{\includegraphics[width=.25\textwidth]{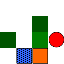}}};

\node[] (ht) at (0,-3.5) {$\underbrace{h_1}_{(0.4,\ 13.1)}$};
\node[] (ht1) at (5,-3.5) {$\underbrace{h_2}_{(5.9,\ -11.6)}$}; 
\node[] (dh) at (2.5,-5) {$dh ={(5.5,\ -24.8)} \approx 2\cdot dh_{{\color{down} down}} + 3\cdot dh_{{\color{right} right}}$};
    \draw[->,thick] (im0.south) -- (ht.north)
    node[midway,fill=white] {Encoder};
    \draw[<-,thick] (ht1) -- (im1.south)
    node[midway,fill=white] {Encoder};
    
    \draw[->,thick] (ht.east) -- (dh.north);
     \draw[->,thick] (ht1.west) -- (dh.north);
    
\end{tikzpicture}}
%  = h + \underbrace{dh_{right}}_{(5,\ -5)} + \underbrace{2\cdot dh_{{\color{green} {\color{green} down}}}}_{(-10,\ -10)}